# Review of Metrics to Measure the Stability, Robustness and Resilience of Reinforcement Learning

Laura L. Pullum, DSc
Computer Science and Mathematics Division, Oak Ridge National Laboratory
pullumll@ornl.gov

**Abstract**

Reinforcement learning (RL) has received significant interest in recent years, due primarily to the successes of deep reinforcement learning at solving many challenging tasks such as playing Chess, Go and online computer games. However, with the increasing focus on RL, applications outside of gaming and simulated environments require understanding the robustness, stability and resilience of RL methods. To this end, we characterize the available literature on these three behaviors as they pertain to RL. We classify the quantification approaches used, determine the objectives of the desired behaviors, and provide a decision tree for selecting metrics to quantify the behaviors.

## 1. Introduction

Recent literature on the robustness of machine learning models has focused almost entirely on the robustness of deep neural networks for imaging applications. However, at the time of this study, there are no published surveys on robustness of RL. With RL use increasing, especially in control systems contexts, we pursued this review. Included along with robustness are stability and resilience. Stability is included because the term has been used interchangeably with robustness and resilience is included because the term has been used as a state beyond robustness.

RL involves agents which take actions in an environment and experience at a reward for those actions. The agent is to learn a policy that maximizes the cumulative reward. Formally, consider an agent operating over time $t \in \{1, \ldots, T\}$. At time $t$, the agent is in environment state $s_t$ and produces an action $a_t \in A$. The agent then observes a new state $s_{t+1}$ and receive a reward $r_t \in R$. The set of possible actions $A$ can be discrete or continuous. The goal of reinforcement learning is to find a policy $\pi(a_t|s_t)$ for choosing an action in state $s_t$ to maximize a utility function or (expected return). [252]

$$J(\pi) = \boldsymbol{E}_{s_0,a_0,\dots}\left[\sum_{t=0}^{\infty} \gamma^t r(s_t, a_t)\right]$$

where $0 \le \gamma \le 1$ is a discount factor; $a_t \sim \pi(a_t|s_t)$ is drawn from the policy; and $s_{t+1} \sim P(s_{t+1}|s_t, a_t)$ is generated by the environmental dynamics. The state value function

$$V^{\pi}(s_t) = \boldsymbol{E}_{a_t,s_{t+1},\dots}\left[\sum_{i=0}^{\infty} \gamma^i r(s_{t+i}, a_{t+i})\right]$$

is the expected return by policy *p* from state $s_t$. The state action function

$$Q^{\pi}(s_t, a_t) = \boldsymbol{E}_{s_{t+1},a_{t+1},\dots}\left[\sum_{i=0}^{\infty} \gamma^i r(s_{t+1}, a_{t+1})\right]$$

is the expected return by policy *p* after taking action $a_t$ at state $s_t$. [252]

The objective of this manuscript is to present a systematic review of RL literature to identify metrics to measure the stability, robustness and resilience of RL. We limit RL to general reinforcement learning and not specialized RL, such as inverse RL. We reviewed papers that attempted to measure or otherwise characterize the stability, robustness and resilience of RL, seeking metrics for these behaviors.

We searched computer science and technical literature databases for eligible papers, combining RL, the behavior terms and terms related to measuring, metrics and quantification. The result comprised 16,015 items, which after removal of duplications and extraneous material, a collection of 546 items was established. Through a process of elimination described in full in the paper, we reduced the set to 248 papers. We systematically reviewed those 248 papers, and the results are presented in this analysis.

We classified the papers by behavior (i.e., stability ($n$=76), robustness ($n$=169), and resilience ($n$=3)) and identified the primary domains of application as robotics, network systems, power system control and vehicle/traffic control and navigation. We identified approaches to determining or measuring each behavior individually and those across behaviors. The approaches were categorized as quantitative or theoretical and the quantitative approaches were further classified as being applied internally (e.g., in training) or externally (e.g., performance measures on outputs) the model. Metrics, approaches and objectives were identified for each paper surveyed. The objective indicates to what the metric or approach was intended to be stable, robust or resilient. We close by indicating the need to define stability, robustness and resilience behaviors for RL and identify the quantitative and theoretical approaches to achieve measurement and determination of these behaviors.

There is a rich set of domains (i.e., 53 identified in this survey) in which measurement of RL stability, robustness and resilience have been conducted. The domains ranged from robotics and network systems to sheep herding and fish behavior. The most frequently mentioned domains include robotics, general control and network systems, with numerous papers not specifying a domain. Many papers used Gym [254] and other environments for demonstration. Though the search focused on quantitative measurement of stability, robustness and resilience, theoretical approaches were identified as well. The quantitative approaches were categorized as internal or external, dependent upon where in the model the evaluation was held. Internal measures quantified the performance of the training, where external measures quantified the ultimate performance of the model.

The goal of this systematic review was to identify metrics to measure the stability, robustness and resilience of RL. To initiate the search for this review, we identified keywords and phrases related to reinforcement learning, the behaviors of interest (stability, robustness and resilience) and measurement (shown in Table 1).

**Table 1.** Keywords and Phrases

| Key Phrase | Behavior | Measurement | |
|---|---|---|---|
| reinforcement learning | stability | metric | measure |
| | robust* | index | score |
| | resilien* | quantifier | indicator |

We believe that this is the first comprehensive review of stability, robustness and resilience specifically geared towards RL. The remainder of the paper is organized as follows. Section 2 describes the methods used in this systematic review. Section 3 presents the results of the review. Section 4 discusses the results of the review and introduces a decision tree for metric selection based on the review. Section 5 provides supplementary information.

## 2. Methods

Keywords salient to RL, system behavior and measurement were identified for the research topic and are shown in Table 1. The typical search was of the form

<Key Phrase> + <Behavior> + <Measurement>

with <Key Phrase>, <Behavior> and <Measurement> defined in Table 1. A specific example is

“reinforcement learning” AND robust* AND (“metric” OR “measure” OR “index” OR “score” OR “quantifier” OR “indicator”)

Multiple searches were conducted in bibliographic databases covering the broad areas of computer science, physical and biological sciences and engineering. See Table 12 for a list of information sources used in this study. No restrictions were placed on the publication’s date or language. Journal articles, books, books in a series, book sections or chapters, edited books, theses and dissertations, conference papers, and technical reports containing the keywords and phrases were included in the search. The publication date of search results returned are bound by the dates of coverage of each database and the date on which the search was performed, however all searches were completed by October 31, 2020. The range of dates for the documents ultimately included in the review was 2002-2020.

The databases queried resulted in 16,015 citations being collected. Irrelevant citations were also unwittingly retrieved. We removed extraneous studies resulting in a collection of 699 publications. Further, removing duplicate papers resulted in 580 publications. Citations for “Full Conference Proceedings” were removed if relevant paper(s) within the associated conference were otherwise collected, resulting in 546 publications. Further refinement excluded publications that were not on RL, that were not on the searched behavior or those that had no metrics or theoretical content, resulting in 248 documents. As a result, we systematically reviewed 248 papers and the results are presented in this analysis. See Figures 10, 11, and 12 for graphic summaries of the data reduction methodology for stability, robustness and resilience behaviors, respectively. See Table 13 for a tabular summary of the data reduction. In addition, see Checklist 1 for the PRISMA (Preferred Reporting Items for Systematic Reviews and Meta-Analyses) guidelines used for the evaluation of the papers.

The 248 papers that made it through the screening process were grouped by the behavior searched, that is Stability, Robustness, and Resilience. We also identified those papers on one behavior that mention one or both of the other behaviors. Some papers that mentioned other behaviors did so interchangeably. For instance, stability and robustness were used interchangeably in several papers, which can lead to some of the confusion that exists in the definitions of these behaviors. The primary domains of application were identified and categorized as robotics, network systems, general control systems and Gym [254] and other environments. We also identified those publications that mentioned the RL policy.

The primary focus of the paper is to identify approaches to determining or measuring each behavior. Of course, most of the publications focused on quantitative approaches because of the search terms used. The ones that use a theoretical approach provide additional insight into the behavior determination problem. The quantitative approaches were further classified as being applied internal (e.g., in training) or external (e.g., performance measures on outputs) the model. Metrics, approaches and objectives were identified for each paper surveyed (see Figure 1). The objective indicates to what the metric or approach was intended to be stable, robust or resilient.

There is little agreement on the definitions of stability, robustness and resilience in the literature. In fact, there are few distinct definitions of these behaviors. For this review, we use the following definitions.

*Stability* is a property of the learning algorithm (that is, a small change in the training set results in a similar model) and refers to the ranking of the variance of a model [253]. For example, if we use variance of the loss function over all datasets as a performance measure and test a set of models. The smallest loss indicates the more stable model. Given this definition, stability analysis is the application of sensitivity analysis to machine learning.

*Robustness*, when used with respect to computer software, refers to an operating system or other program that performs well not only under ordinary conditions but also under unusual conditions that stress its designers' assumptions (http://www.linfo.org/robust.html). Robustness then is a property of the model and is measured by the, e.g., loss over all datasets (as opposed to the variance of the loss).

Throughout the literature, *resilience* has been used interchangeably with robustness, however, it is used most often with production machine learning systems to indicate robustness to different data sets and different data added to the data set.

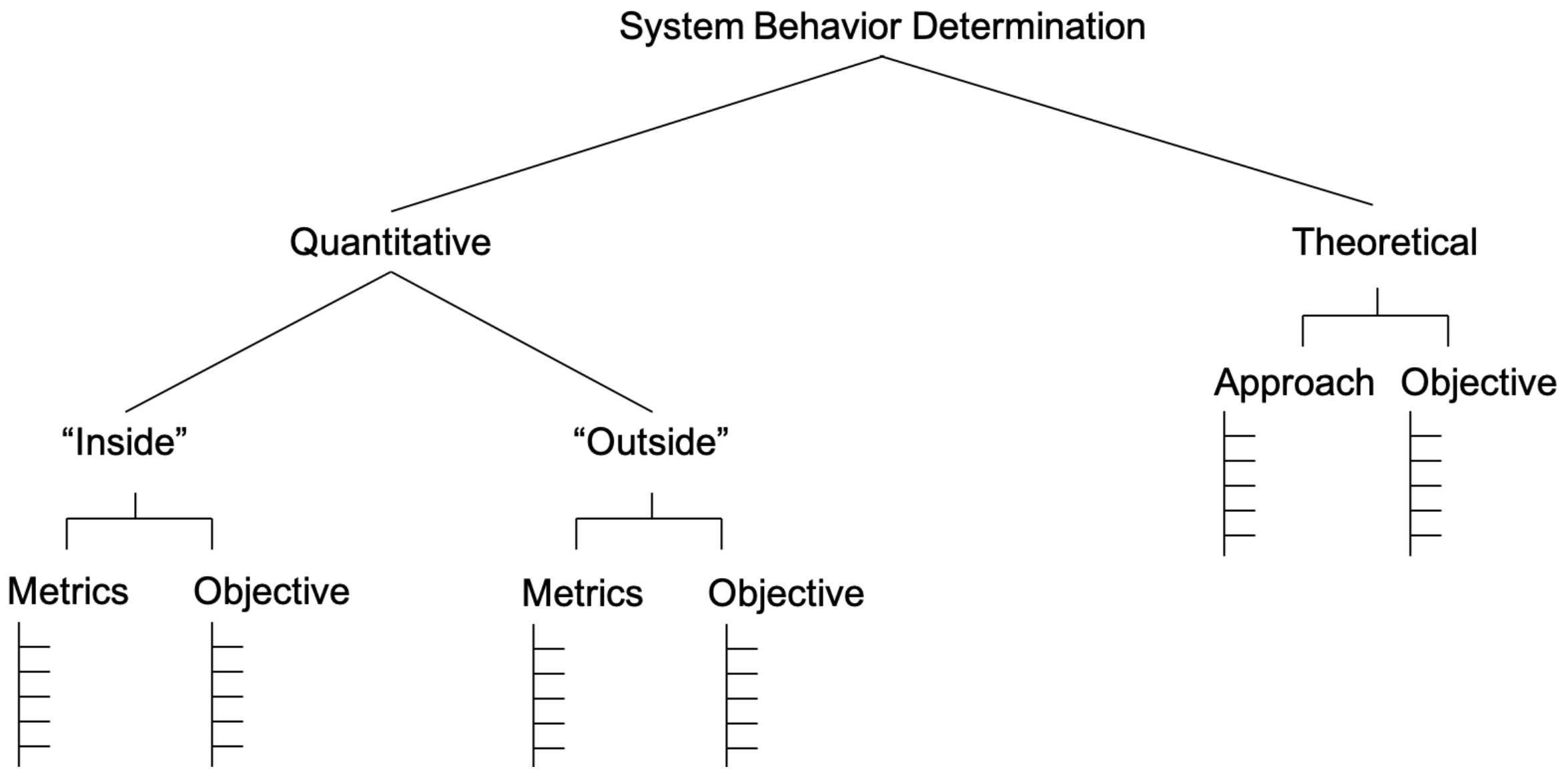

**Figure 1.** Categorization and resulting metrics, approaches and objectives

## 3. Results and Analysis

The publications were categorized by behavior (see Table 2): Stability (*n*=76), Robustness (*n*=169), and Resilience (*n*=3). Papers on one behavior often mention the other behaviors, especially Stability and Robustness (see Figure 2). Resilience is mentioned in 5 Stability papers and in 11 Robustness papers. Robustness is mentioned in 50 Stability papers and in 1 Resilience paper. Stability is mentioned in 104 Robustness papers and in all (3) Resilience papers.

**Table 2.** Citations categorized by behavior

| Behavior | Citations | Total |
|---|---|---|
| Stability | [4-80] | 76 |
| Robustness | [81-250] | 169 |
| Resilience | [1-3] | 3 |
| | Total | 248 |

Given the recent explosion of papers on robustness of neural networks to adversarial attacks, one might expect it to be a cornerstone of the robustness papers reviewed herein. The term "adversarial" is mentioned in a quarter ($n$=61, $N$=248) of the papers reviewed (see Figure 3). That is, 1 Resilience paper, 56 Robustness papers and 4 Stability papers mention "adversarial". Some papers on one behavior used one of the other behaviors interchangeably, notably stability and robustness, specifically [91, 93, 105, 145, 146, 179, 194, 225, and 237] and generally in several other articles.

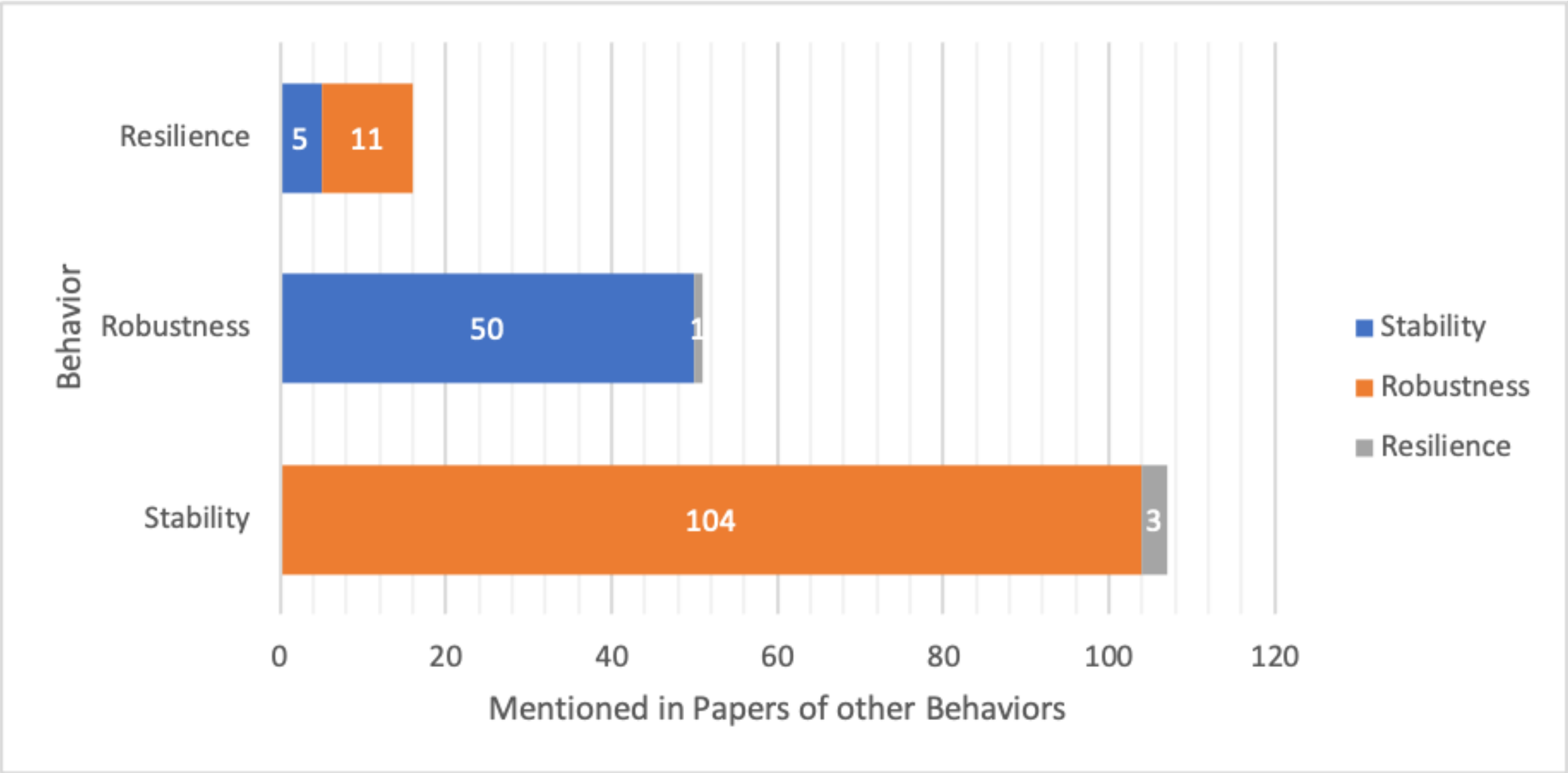

**Figure 2.** Stability, robustness and resilience papers were mentioned in papers on other behaviors. For example, Robustness is mentioned in 104 Robustness papers and in all 3 Resilience papers.

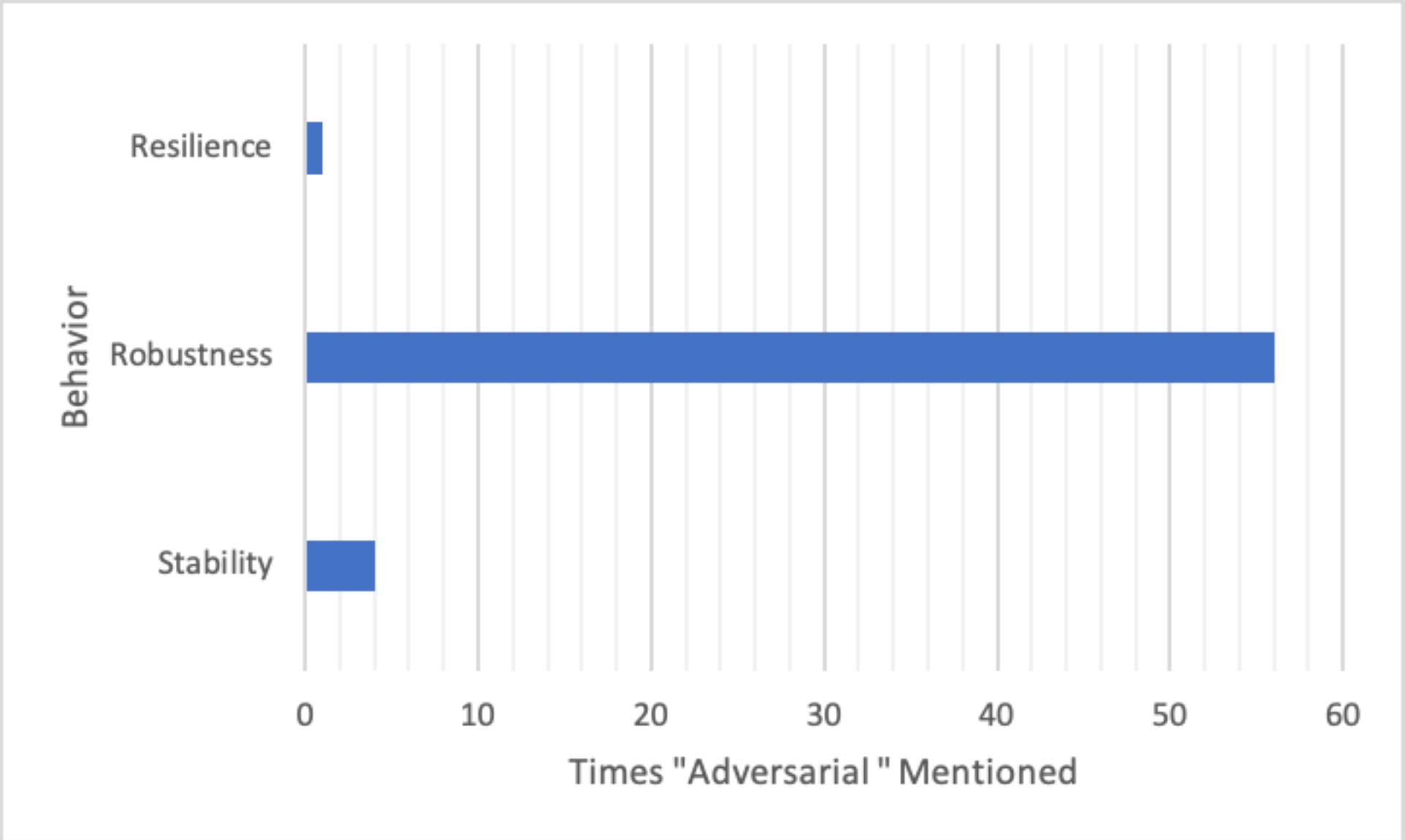

**Figure 3.** Number of papers in which "adversarial" is mentioned, categorized by behavior. For example, 56 Robustness papers mention "adversarial".

### 3.1 Application Domains

The publications' application domains are provided in Table 3 and illustrated in Figure 4. The primary domains were robotics with 16.4% (*n*=44) of the total citations (*N*=268), followed by network systems and general control (each with 7.8%, *n*=21), with 9.3% (*n*=25) using Gym or other environments as their experiment domain. Just as many (*n*=25, 9.3%) did not specify a domain. These top 5 (of 53) domains comprised over 50% (52.9%, *n*=136) of the citations. Most (52.8%, *n*=28) of the domains (*n*=53) had a single citation each.

**Table 3.** Citations categorized by application domain

| Domain | Behavior(s) | Citations | Total |
|---|---|---|---|
| Robotics | Stability, Robustness | [9, 12, 16, 21, 27, 31, 44, 45, 46, 51, 56, 60, 70, 77, 78], [82, 86, 103, 109, 116, 118, 122, 125, 128, 133, 134, 140, 142, 149, 151, 158, 162, 164, 178, 190, 208, 211, 212, 217, 221, 224, 236, 242, 249] | 44 |
| Gym and other environments | Stability, Robustness | [8], [87, 101, 146, 182, 185, 191, 192, 195, 199, 200, 208, 109, 213, 225-229, 231, 233, 240, 245, 246, 250] | 25 |
| General, non-specified | Robustness | [132, 136, 145, 150, 171, 173, 174, 181, 184, 185, 195, 197, 198, 200, 205, 215, 216, 219, 223, 225, 229, 240, 247, 248, 250] | 25 |
| Network Systems | Stability, Robustness, Resilience | [6, 10, 11, 14, 32, 42, 66, 73, 77], [81, 84, 89, 96, 98, 104, 113, 115, 127, 203, 214], [3] | 21 |
| Control, not otherwise noted | Stability, Robustness | [64, 68, 69, 71, 74, 76, 79], [135, 137, 147, 163, 175, 188, 189, 194, 210, 218, 228, 230, 233, 244] | 21 |
| Vehicle/Traffic Control and Navigation, Collision Avoidance | Stability, Robustness | [16, 20, 38, 47, 49, 58, 65], [83, 100, 110, 111, 114, 123, 166, 177, 195, 226] | 17 |
| Games/Game Systems | Stability, Robustness | [24, 30, 33, 35, 37, 52], [156, 165, 168, 193, 201, 204, 220, 234, 239, 250] | 16 |
| Power System Control | Stability, Robustness | [4, 10, 15, 22, 23, 36, 43, 48, 50, 59, 67], [105, 107, 117, 172] | 15 |
| Image Analysis | Stability, Robustness | [72], [88, 93, 94, 99, 108, 148, 155, 235] | 9 |
| Nonlinear Dynamic Systems/Processes | Stability | [5, 25, 41, 54, 61, 62] | 6 |
| Continuous Control | Robustness | [157, 191, 231, 237, 245] | 5 |

**Table 3.** Citations categorized by application domain

| Domain | Behavior(s) | Citations | Total |
|---|---|---|---|
| Multi-agent Systems | Stability, Robustness | [18, 29, 52], [241] | 4 |
| Domain Agnostic | Stability | [7, 28, 40] | 3 |
| Manufacturing Systems | Stability, Robustness | [10, 63], [95] | 3 |
| Ride-share Dispatching, Delivery | Stability, Robustness | [19], [97, 160] | 3 |
| Medical - System Control, Medication Level/Control | Robustness | [112, 144, 187] | 3 |
| Autonomous Systems | Robustness | [143, 196, 207] | 3 |
| Security & Cyber Defense, Spammer Detection | Robustness | [138, 159, 179] | 3 |
| Uncertain Non-linear Systems | Robustness | [106, 121] | 2 |
| Conversation and Speech | Robustness | [119, 124] | 2 |
| Text Analysis, Machine Translation | Stability, Robustness | [13], [85] | 2 |
| Multi-armed Bandit | Stability, Robustness | [74], [120] | 2 |
| Feedback Controller | Stability, Robustness | [73], [129] | 2 |
| Information Retrieval (search) | Stability, Robustness | [17], [161] | 2 |
| Economics - Quantitative Investment, Trading Systems, Markets | Robustness | [102, 130, 206] | 2 |
| Video Presentation Quality | Stability | [26] | 1 |
| Brain-Machine Interface (BMI) Controller | Stability | [34] | 1 |
| Skill Acquisition | Stability | [53] | 1 |

**Table 3.** Citations categorized by application domain

| Domain | Behavior(s) | Citations | Total |
|---|---|---|---|
| Technical Process Feedback | Stability | [55] | 1 |
| Linear Quadratic Output Tracking | Stability | [39] | 1 |
| Decentralized Control | Robustness | [90] | 1 |
| Policy-selection Races | Robustness | [126] | 1 |
| Complex Adaptive Systems | Resilience | [1] | 1 |
| Emission Control | Stability | [57] | 1 |
| Flight Simulator Training | Resilience | [2] | 1 |
| Markov Decision Process | Stability | [75] | 1 |
| Cellular Communications | Robustness | [167] | 1 |
| Fish Behavior | Robustness | [141] | 1 |
| Space Telescope | Robustness | [131] | 1 |
| Brain Modeling | Robustness | [139] | 1 |
| Sensors | Robustness | [152] | 1 |
| Unsupervised Goal Exploration | Robustness | [153] | 1 |
| Directed Graph Embedding | Robustness | [154] | 1 |
| HVAC Control | Robustness | [169] | 1 |
| Train | Robustness | [170] | 1 |
| Teamwork | Robustness | [176] | 1 |
| Causal Discovery | Robustness | [180] | 1 |
| Turbo Fan Engines | Robustness | [183] | 1 |
| Sheep Herding | Robustness | [186] | 1 |
| Van de Pol Oscillator | Robustness | [222] | 1 |
| Single Episode Transfer | Robustness | [232] | 1 |
| Multi-Task Batch | Robustness | [238] | 1 |
| Spectral Efficiency | Robustness | [243] | 1 |
| | | Total Citations | 268 |
| The categories are not mutually exclusive. | | Total Domains | 53 |

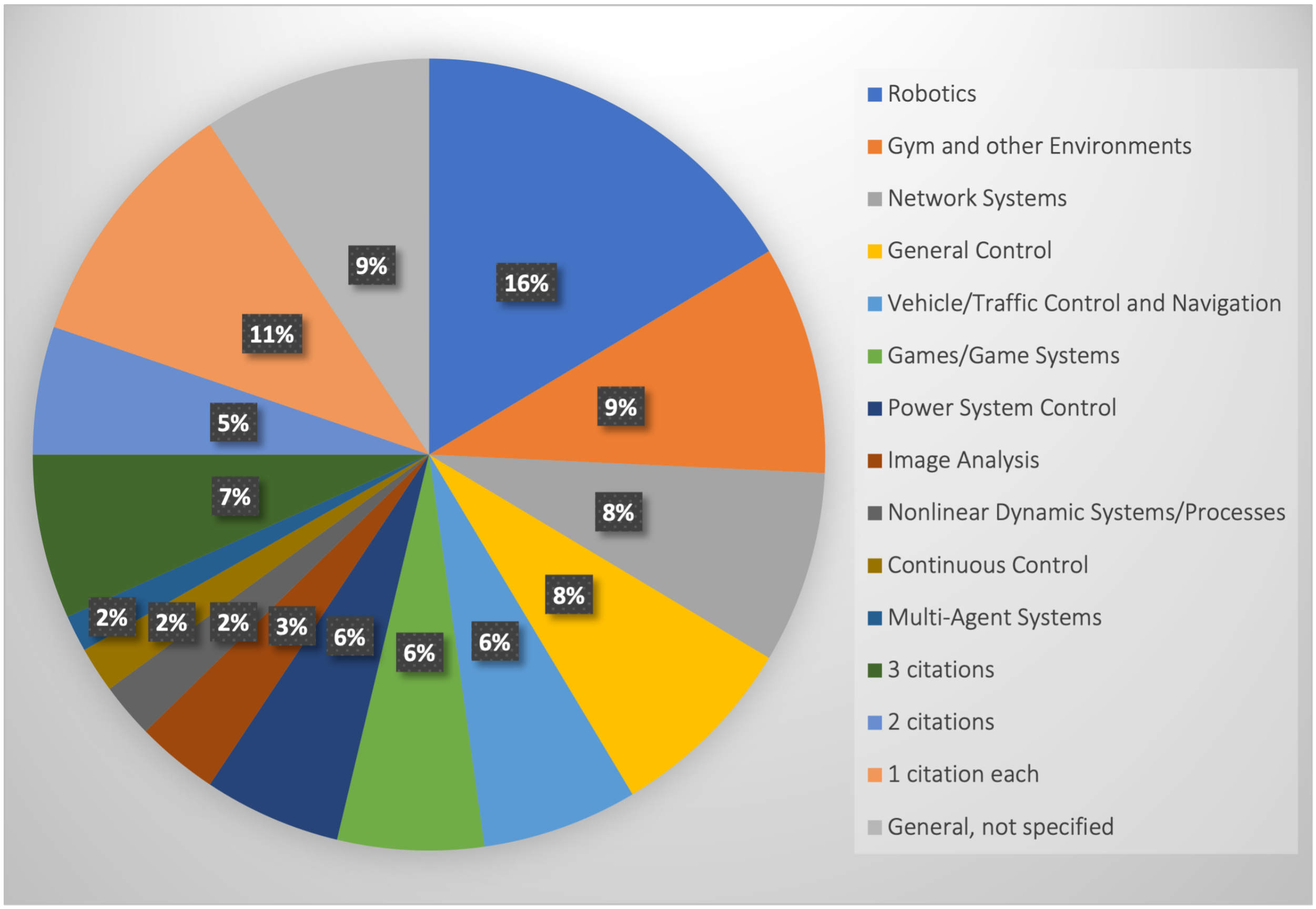

**Figure 4.** Application Domain Categories

### 3.2 Reinforcement Learning Policies

The types of RL policies mentioned in the articles are provided in Table 4. Most documents did not identify the policy used. If a behavior is not provided in Table 4 for a particular policy, that policy was not discussed in the papers on that behavior. Of the 21 types of policies mentioned, the top 4 – Actor-Critic ($n$=18), Q-learning ($n$=16), Proximal Policy Optimization (PPO) ($n$=8) and Adaptive Critic Design ($n$=5) comprise 72.3% of the total citations that included policy ($n$=65).

**Table 4.** Citations categorized by the reinforcement learning policy used

| Policy | Behavior | Citations | Subtotal | Total |
|---|---|---|---|---|
| Actor-Critic | Stability | [5, 10, 46, 47, 49, 51, 65, 68, 69, 77, 79] | 11 | 18 |
| | Robustness | [131, 155, 180, 185, 191, 192, 222] | 7 | |
| Q-Learning | Stability | [18, 19, 39, 66, 72-75] | 6 | 16 |
| | Robustness | [92, 94, 107, 109, 113, 130, 138, 223, 236] | 9 | |
| | Resilience | [3] | 1 | |
| PPO | Robustness | [143, 147, 149, 158, 161, 162, 166, 208] | 8 | 8 |
| Adaptive Critic Design | Stability | [43, 54, 61] | 3 | 5 |
| | Robustness | [110, 125] | 2 | |
| Deep Deterministic Policy Gradient (DDPG) | Robustness | [162, 167] | 2 | 2 |
| 4 variational Model-based Policy Optimization | Robustness | [181] | 1 | 1 |
| Advantage Actor Critic (A2C) | Robustness | [143] | 1 | 1 |
| Ad-hoc On-demand Distance Vector (AODV) | Robustness | [123] | 1 | 1 |
| Active Tracking Target Network (ATTN) and Anytime Reduced Value Iteration (ARVI) | Robustness | [177] | 1 | 1 |
| Constant Feedback to Control Policy | Robustness | [106] | 1 | 1 |

**Table 4.** Citations categorized by the reinforcement learning policy used

| **Policy** | **Behavior** | **Citations** | **Subtotal** | **Total** |
|---|---|---|---|---|
| Constraint-controlled PPO (CPPO) | Robustness | [164] | 1 | 1 |
| Data-regulated Actor-Critic (DrAC) | Robustness | [171] | 1 | 1 |
| Deep Deterministic Policy Gradient | Stability | [70] | 1 | 1 |
| Generalized Advantage Estimation | Robustness | [158] | 1 | 1 |
| Goal Policy | Robustness | [153] | 1 | 1 |
| Gradient | Robustness | [179] | 1 | 1 |
| Graph-based Policy Learning | Robustness | [176] | 1 | 1 |
| Lyapunov-based Actor-Critic (CLAC) | Robustness | [183] | 1 | 1 |
| MOOSE (MOdel-based Offline policy Search with Ensembles) | Robustness | [151] | 1 | 1 |
| Natural Stochastic Policies | Robustness | [184] | 1 | 1 |
| Student $t$ Policy | Robustness | [146] | 1 | 1 |
| | | Total Citations including Policy | | 65 |
| | | Total Policies | | 21 |

## 3.3 Approach to Determining or Measuring Behavior

The publications' approaches to determining or measuring each behavior are categorized as either quantitative or theoretical (Table 5). Most of the publications focused on quantitative approaches ($n$=205, 82.0%), which is understandable given that the search focused on quantifying the behaviors. For publications on the stability behavior, there was an almost even split between quantitative (*n*=42) and theoretical (*n*=43) approaches. However, publications on the robustness behavior were primarily focused on quantitative approaches. All (3) resilience publications applied quantitative approaches.

**Table 5.** Citations categorized by approach to determining or measuring behavior

| Approach | Behavior | Citations | Total |
|---|---|---|---|
| Quantitative | Stability | [4, 6, 7, 8, 9, 10, 11, 12, 13, 14, 15, 17, 18, 19, 20, 21, 23, 26, 27, 28, 29, 30, 31, 32, 34, 36, 38, 42, 43, 45, 50, 51, 53, 58, 59, 66, 70, 72, 73, 75, 78, 80] | 42 |
| | Robustness | [81-89, 92-104, 106-120, 122-196, 198-201, 203-222, 224-227, 229, 231-247, 249-250] | 160 |
| | Resilience | [1, 2, 3] | 3 |
| | | Total Quantitative | 205 |
| Theoretical | Stability | [4, 5, 10, 13, 15, 16, 20, 22, 24, 25, 27, 33, 35, 36, 37, 39, 40, 41, 43, 44, 46, 47, 48, 49, 52, 54, 55, 56, 57, 60, 61, 62, 63, 64, 65, 67-69, 71, 74, 76-79] | 44 |
| | Robustness | [90, 91, 97, 105, 116, 121, 135, 145, 150, 165, 184, 185, 194, 195, 197, 200-202, 205, 206, 215, 216, 222, 223, 225-230, 233, 239, 240, 248, 250] | 35 |
| | Resilience | -- | 0 |
| | | Total Theoretical | 79 |

If a document covered both quantitative and theoretical approaches, it was placed in both categories.

#### *3.3.1 Types of Quantitative Approaches*

Next, we further categorize the quantitative approaches into whether they are focused internal or external to the model (see Table 6). Internal quantitative approaches measure aspects within the model, e.g., its training and associated measures such as the value of rewards over time or the number of episodes until convergence. External quantitative approaches measure performance-related aspects of the model, e.g., variations in accuracy or throughput. Most (*n*=141, 63.2%) of the quantitative approaches were categorized as performance-related, or external, measures. Of these, most (*n*=103) were on robustness, with stability (*n*=36) next. The 3 papers on resilience focused on performance-related quantitative measures. Robustness also led the internal approaches (*n*=68) with stability following (*n*=14). These are primarily due to the large number of robustness papers (*n*=170) and paucity of resilience papers (*n*=3) overall. Of the robustness papers, 40.0% (*n*=68) contained internal quantitative measures and 60.6% contained external quantitative measures. For stability, these values are 18.2% and 46.8%, respectively.

**Table 6.** Quantitative approaches categorized by internal or external measures

| Quantitative Approach | Behavior | Citations | Total |
|---|---|---|---|
| Internal | Stability | [7, 8, 9, 11, 13, 19, 20, 27, 29, 30, 43, 72, 75, 78] | 14 |
| | Robustness | [83, 84, 92, 94, 97, 101-103, 109, 114, 122, 130, 132, 134, 138, 143, 146, 149-151, 155-157, 163-164, 166, 174-176, 181-182, 185, 187-189, 191-193, 195, 198-200, 204-205, 208-209, 211-213, 215, 217, 220, 226, 229, 231-239, 241, 244-247, 250] | 69 |
| | Resilience | -- | 0 |
| | | **Total Internal Measures** | **83** |
| External | Stability | [6, 9, 10, 12, 13, 14, 15, 17, 18, 19, 20, 21, 23, 26, 27, 28, 31, 32, 34, 36, 38, 42, 45, 50, 51, 53, 58, 59, 66, 68, 70, 73, 75, 78, 80] | 36 |
| | Robustness | [81-83, 85-89, 93, 95-100, 102, 104, 106-120, 123-129, 131, 133, 135-145, 147-148, 152-154, 158-162, 165-173, 177-181, 183-184, 186, 190, 194-196, 201, 203-204, 206-207, 210, 214, 216, 218-219, 221-222, 224-227, 240, 242-243, 249] | 103 |
| | Resilience | [1, 2, 3] | 3 |
| | | **Total Performance-related Measures** | **142** |

If there were both internal and external quantitative approaches in a document, the document was placed in both categories.

### *3.3.2 Types of Internal Quantitative Approaches*

Looking at the types of internal quantitative approaches, we see a fairly narrow set of aspects being considered in the papers (see Table 7). These are metrics specifically made to measure stability other than by the variance of the output. They essentially measure variation in training performance. The vast majority ($n$=75, 88.2%) of the internal quantitative approaches calculate reward- or score-based metrics. Other types of internal quantitative approaches include 2 each of policy entropy, variations in control strategy approximation weights, and the convergence rate, and 1 each of policy weight, calculation of the Lyapunov stability criteria and calculation of the Wasserstein function lower bound. The term *convergence*, in RL context, refers to the stability of the learning process (and the underlying model) over time [11].

**Table 7.** Internal Quantitative approaches categorized by metric

| **Internal Quantitative Metric** | **Behavior** | **Citations** | **Total** |
|---|---|---|---|
| Reward or Score – magnitude, mean/ variance, variation in average reward, time to threshold, episode duration | Stability, Robustness | [7, 8, 9, 13, 30, 72, 75, 78] [83, 84, 92, 94, 97, 101-103, 109, 114, 122, 130, 132, 134, 138, 143, 146, 149-151, 155-157, 163-164, 166, 174-176, 181-182, 185, 188-189, 191-193, 195, 198-200, 204, 208-209, 211-213, 215, 217, 220, 226, 229, 231-239, 241, 244-247, 250] | 75 |
| Policy entropy | Stability | [11, 19] | 2 |
| Variations in control strategy approximation weights | Stability, Robustness | [20] [120] | 2 |
| Convergence rate | Stability | [27, 29] | 2 |
| Lyapunov stability criteria calculated | Stability | [43] | 1 |
| Policy weight | Robustness | [231] | 1 |
| Regret | Robustness | [187] | 1 |
| Wasserstein function bounds calculated | Robustness | [205] | 1 |
| | | **Total** | **85** |

If a document had metrics that fell in more than one category, the document was placed in each of the categories.

#### *3.3.3 Types of External Quantitative Approaches*

The external or performance-based quantitative approaches to measuring the behaviors primarily ($n$=39) used deviations or variation in performance-related metrics other than precision, accuracy or recall (see Table 8 and Figure 5). The next category ($n$=28) of quantitative metrics used error, failure and success rates. Statistics on the performance of the tracking or estimation error follows with $n$=23 papers. Papers in the network domain used network-related metrics ($n$=15) to measure the behavior. Statistics on precision, accuracy and recall ($n$=12) followed. Five papers used variance in loss or regret estimation, 3 papers used game-related performance measures to quantify behavior and 2 papers each used bounds on or the size of the stability region and terminal wealth and inventory. Eighteen (18) additional different types of external quantitative metrics categories were represented by a single paper each.

#### *3.3.4 Quantitative Approach Objectives*

An additional aspect reviewed was to what action or event were the quantitative approaches attempting to be stable, robust or resilient. We call this the *<behavior> objective*. The *<behavior>* objective category (see Table 9) with the highest number of citations was geared toward handling changes in the operational environment or a dynamic environment or network ($n$=41). Papers that did not specifically state their objective comprised the next most populous category ($n$=35). The objective of handling uncertainty and disturbances in the environment also contained $n$=35 papers. The remaining objectives included input variation/perturbations ($n$=20); differences between training and test or operational environments ($n$=19); differences or uncertainties in model parameters ($n$=16); adversarial attack ($n$=14); different domains, environments or settings ($n$=8); errors or failures in operational environment ($n$=5); differences in training data sets or initializations ($n$=5); high variability ($n$=2) and one paper each in systematic pressure, spamming, incomplete data, and unknown control coefficients.

**Table 8.** External Quantitative approaches categorized by metric

| External Quantitative Metric | Behavior | Citations | Total |
|---|---|---|---|
| Deviations/variation in other (than precision, accuracy and recall) performance-related metrics | Stability,<br>Robustness,<br>Resilience | [9, 13, 15, 19, 20, 26, 27, 36, 38, 45, 50, 51, 58, 59, 68, 80]<br>[85, 95, 97, 106, 107, 110, 111, 119, 126, 137-139, 141, 142, 152, 169, 178, 179, 181, 183, 184, 195]<br>[2] | 39 |
| Error and failure rates/success rate | Stability,<br>Robustness | [12]<br>[86, 88, 100, 104, 112, 124, 129, 131, 133, 136, 140, 143, 153, 158, 160, 162, 166-168, 186, 204, 207, 221, 222, 225, 242, 249] | 28 |
| Performance of tracking/trajectories estimation error; mean absolute deviation, mean square error, mean absolute percentage error, margins and magnitude of correlation coefficient | Stability,<br>Robustness | [10, 18, 23, 28, 31]<br>[87, 88, 93, 96, 106, 116, 125, 144, 145, 147, 148, 158, 173, 186, 201, 224, 226, 240] | 23 |
| Network-related timing/delay, path and link metrics, connectivity, delivery ratio, routing loops, path optimality, visitation distribution, structural Hamming distance, Small base station-serving ratio, sum-rate and 5$^{th}$ percentile rate | Stability,<br>Robustness | [6, 14, 32, 42, 53, 66, 73]<br>[83, 89, 123, 127, 170, 180, 203, 214] | 15 |
| Mean/average and variation in accuracy, precision and recall, area under the receiver operating characteristic (ROC) curve (AUC) | Stability,<br>Robustness,<br>Resilience | [17, 34]<br>[93, 99, 102, 108, 113, 154, 159, 161, 168, 227]<br>[3] | 12 |
| Variance of the estimation of loss, regret | Robustness | [118, 159, 187, 216, 224] | 5 |

**Table 8.** External Quantitative approaches categorized by metric

| External Quantitative Metric | Behavior | Citations | Total |
|---|---|---|---|
| Game-related performance - scores of game playing, percent wins, exploitability | Robustness | [109, 120, 165] | 3 |
| Size of the stability region; bounds | Stability,<br>Robustness | [21]<br>[135] | 2 |
| Terminal wealth, terminal inventory, cost, Sharpe ratio | Robustness | [206, 218] | 2 |
| Average proportion of failed eavesdropping attempts and of jammed red-force nodes; Average throughput | Robustness | [81] | 1 |
| Hours of operation with some maintenance | Robustness | [82] | 1 |
| Response time, energy consumption, and execution time | Robustness | [98] | 1 |
| Spectral efficiency | Robustness | [243] | 1 |
| Time headway (sec) | Robustness | [210] | 1 |
| Covariance analysis as a metric | Robustness | [114] | 1 |
| Mutual information | Robustness | [177] | 1 |
| Fidelity | Robustness | [219] | 1 |
| Expected rank and Robust Measurements metric | Robustness | [115] | 1 |
| Singular value decomposition (SVD)-based controllability measure | Robustness | [117] | 1 |
| Time to find goal/destination | Robustness | [128] | 1 |
| Number of adversarial actions required to cause error | Resilience | [1] | 1 |
| Normalized Energy Stability Margin (NESM) | Stability | [70] | 1 |
| Voltage violation rate, active power loss | Robustness | [172] | 1 |
| Jensen-Shannon divergence (JSD) | Robustness | [171] | 1 |
| Number of successful steps | Robustness | [190] | 1 |
| Blood glucose responses, Insulin concentration | Robustness | [194] | 1 |

**Table 8.** External Quantitative approaches categorized by metric

| External Quantitative Metric | Behavior | Citations | Total |
|---|---|---|---|
| Likelihood (Mahalanobis Distance) | Robustness | [196] | 1 |
| | | **Total** | **147** |

If a paper contained quantitative approaches belonging to multiple categories, it was placed in each of the relevant categories.

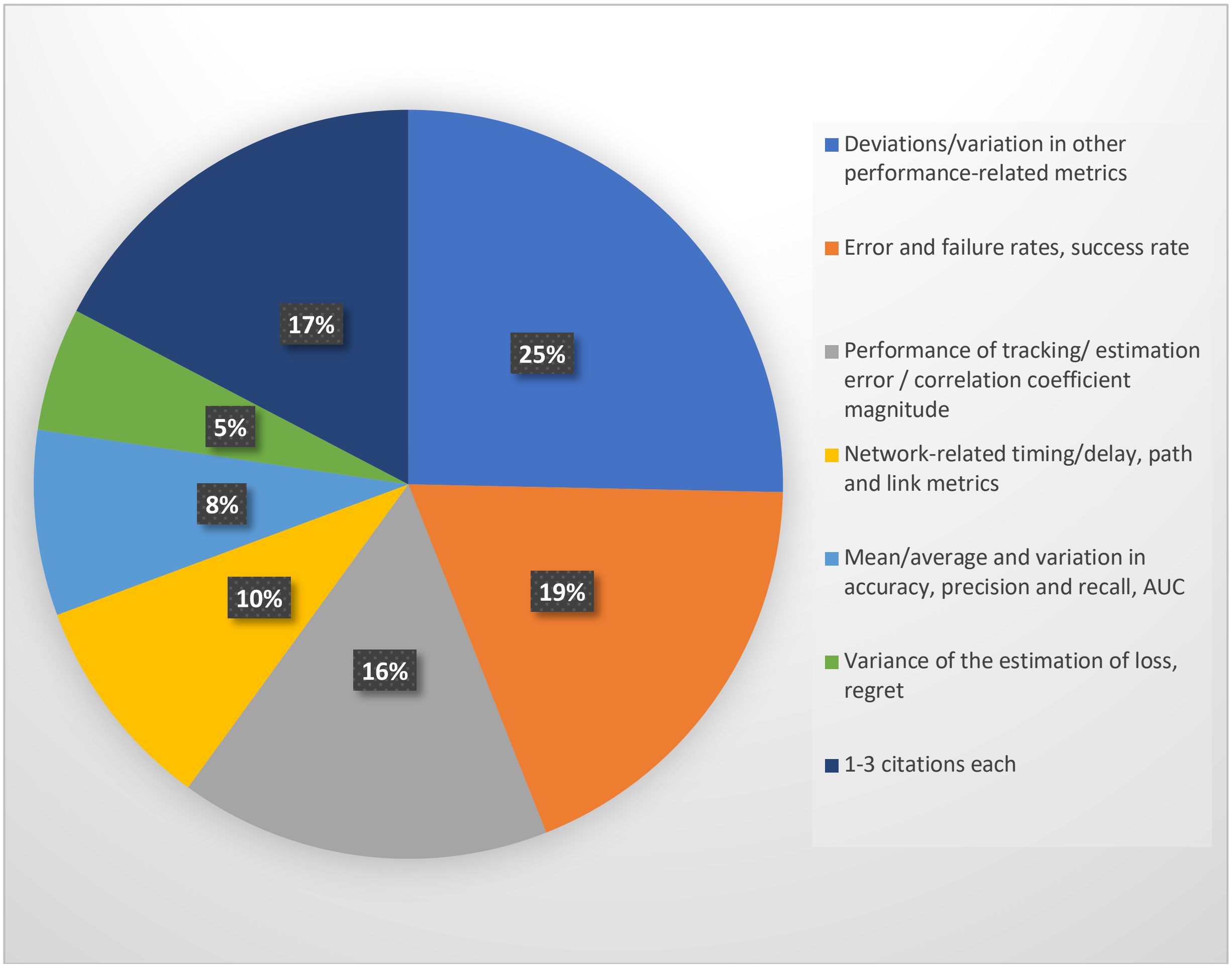

**Figure 5.** External quantitative metrics

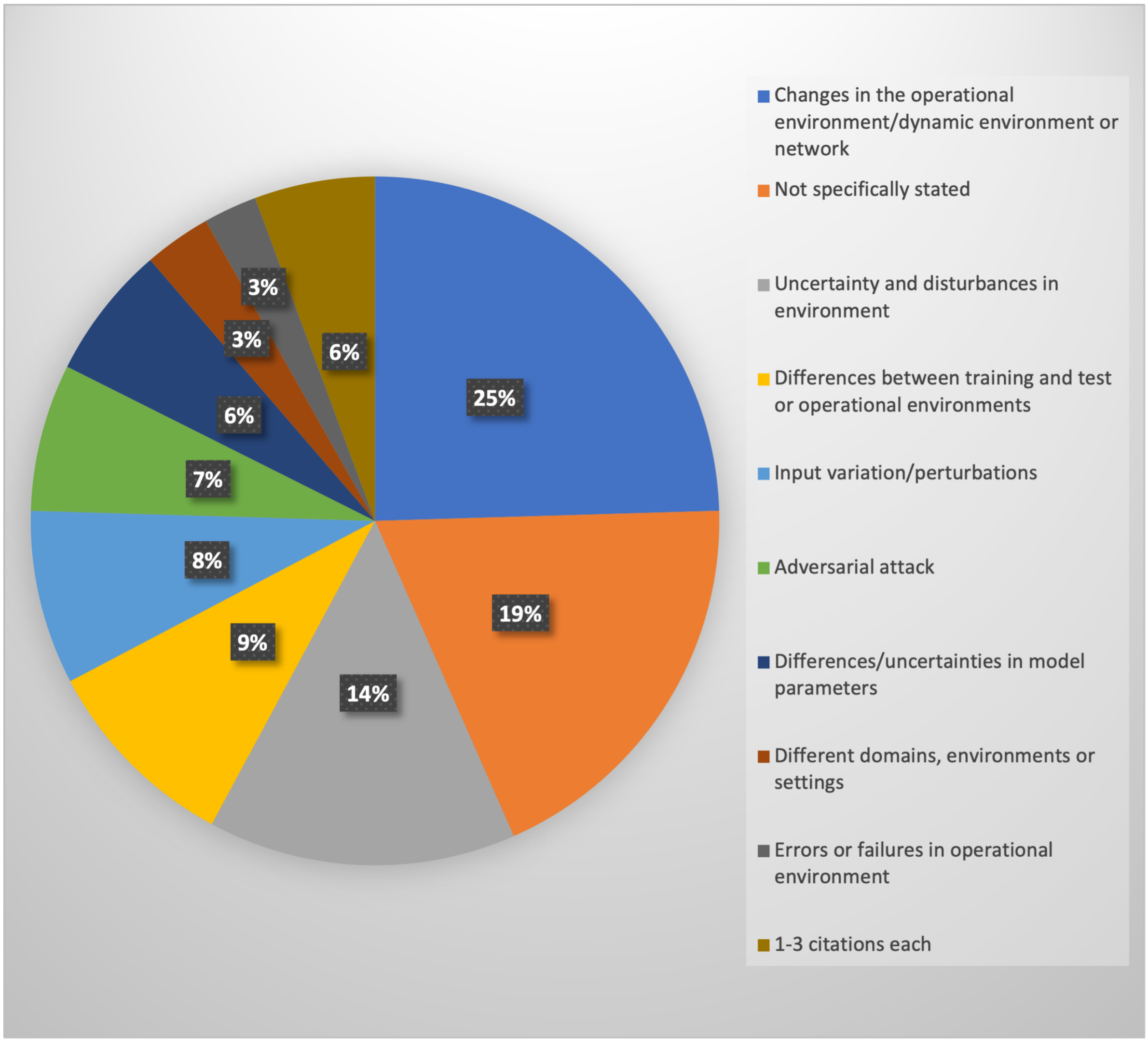

**Figure 6.** Quantitative <*behavior*> objectives

### *3.3.5 Types of Theoretical Approaches*

A majority of the types of theoretical approaches in the papers reviewed were based on Lyapunov theory ($n$=50, 61.0%) (Table 10). The next highest types of theoretical approaches used are convergence to Nash equilibrium ($n$=10) and value-based guarantees such as error and output deviation bounds ($n$=8). Of the remainder, 3 papers used the Wasserstein distance to explore stability, 3 papers proved the methods were doubly robust, 2 papers proved the methods exhibited Lipschitz continuity, and stochastic stability theory to prove stability, stability guarantees, policy-based guarantees, regret bounds, minimization of the Jacobian on input, and per-episode Bellman-error regret guarantees/bounds were used by a single paper each to establish stability of the RL methods discussed.

**Table 9.** Quantitative <*behavior*> objectives

| **<*behavior*> Objective** | **Behavior(s)** | **Citations** | **Total** |
|---|---|---|---|
| Changes in the operational environment/dynamic environment or network; distribution shift | Stability<br>Robustness<br>Resilience | [26, 27, 32, 45, 50, 66, 70, 73]<br>[85, 86, 89, 93, 94, 96, 98, 99, 102, 107, 108, 111, 112, 114, 117, 118, 122, 123, 128, 131, 142, 154, 167, 171, 172, 177, 187, 194, 207, 214, 220, 243]<br>[3] | 41 |
| Not specifically stated | Stability<br>Robustness | [6, 8, 9, 10, 11, 13, 14, 15, 17, 18, 20, 21, 23, 28, 31, 38, 43, 51, 58, 59, 78]<br>[113, 116, 125, 127, 130, 134, 165, 184, 198, 209, 216, 227, 240, 247] | 35 |
| Uncertainty and disturbances in environment, e.g., noisy sensor data, measurement noise, distractors, nuisances | Robustness | [133, 135, 138, 140, 141, 147, 148, 152, 153, 162, 169, 173, 175, 178, 186, 188, 190, 191, 193, 195, 196, 199, 208, 211, 217, 218, 219, 222, 224, 225, 234-236, 244, 249] | 35 |
| Input variation/perturbations, outliers | Stability<br>Robustness | [7, 12, 19]<br>[101, 103, 106, 132, 139, 143, 146, 149, 157, 158, 170, 187, 195, 208, 221, 226, 231] | 20 |
| Differences between training and test or operational environments | Stability<br>Robustness | [31, 42]<br>[82-84, 87, 95, 100, 109, 119, 120, 124, 129, 136, 149, 174, 191, 221, 245] | 19 |

**Table 9.** Quantitative <*behavior*> objectives

| **<*behavior*> Objective** | **Behavior(s)** | **Citations** | **Total** |
|---|---|---|---|
| Differences/uncertainties in model [hyper-]parameters, model error | Stability<br>Robustness | [36, 53, 75]<br>[92, 110, 126, 137, 164, 174, 181, 182, 192, 201, 225, 237, 241] | 16 |
| Adversarial attack | Stability<br>Robustness<br><br>Resilience | [29, 30]<br>[81, 97, 156, 159, 163, 166, 204, 206, 210, 212]<br>[1, 2] | 14 |
| Different domains, environments or settings | Stability<br>Robustness | [80]<br>[161, 164, 176, 203, 224, 232, 242] | 8 |
| Errors or failures in operational environment | Robustness | [104, 115, 145, 168, 189] | 5 |
| Differences in training data sets or initializations | Stability<br>Robustness | [34]<br>[88, 151, 213, 238] | 5 |
| High variability | Robustness | [144, 183] | 2 |
| Systematic pressure, e.g., sudden surge of requests | Robustness | [160] | 1 |
| Spamming | Robustness | [179] | 1 |
| Incomplete data | Robustness | [180] | 1 |
| Unknown control coefficients | Stability | [68] | 1 |
| | | **Total** | **204** |

If a technique had multiple robustness objectives, it was placed in each of those objectives.

**Table 10.** Theoretical approaches categorized by specific approach

| Theoretical Approach | Behavior | Citations | Total |
|---|---|---|---|
| Lyapunov stability theory | Stability | [4, 5, 10, 15, 16, 20, 25, 35, 36, 39, 40, 41, 43, 44, 46, 47, 48, 49, 52, 54, 55, 56, 57, 60, 61, 62, 64, 65, 67, 69, 74, 76-79] | 50 |
| | Robustness | [90, 91, 97, 105, 116, 121, 135, 145, 194, 200, 222, 225, 230, 233, 248] | |
| Convergence to Nash equilibrium | Stability | [22, 24, 37, 63, 71] | 10 |
| | Robustness | [165, 197, 206, 215, 230, 239] | |
| Value-based guarantees, error bounds, output deviation bounds | Robustness | [195, 201, 226-228, 230, 233, 250] | 8 |
| Wasserstein distance | Robustness | [185, 201, 205] | 3 |
| Prove double robustness | Robustness | [184, 216, 240] | 3 |
| Prove Lipschitz continuity | Robustness | [201, 229] | 2 |
| Stochastic stability theory | Stability | [33] | 1 |
| Prove stability guarantees | Stability | [68] | 1 |
| Policy-based guarantees | Robustness | [201] | 1 |
| Regret bounds | Robustness | [202] | 1 |
| Minimize Jacobian on input | Robustness | [150] | 1 |
| Per-episode Bellman-error regret guarantees/bounds | Robustness | [223] | 1 |
| | | **Total** | **82** |

If a paper used multiple theoretical approaches, the was placed in each of those categories.

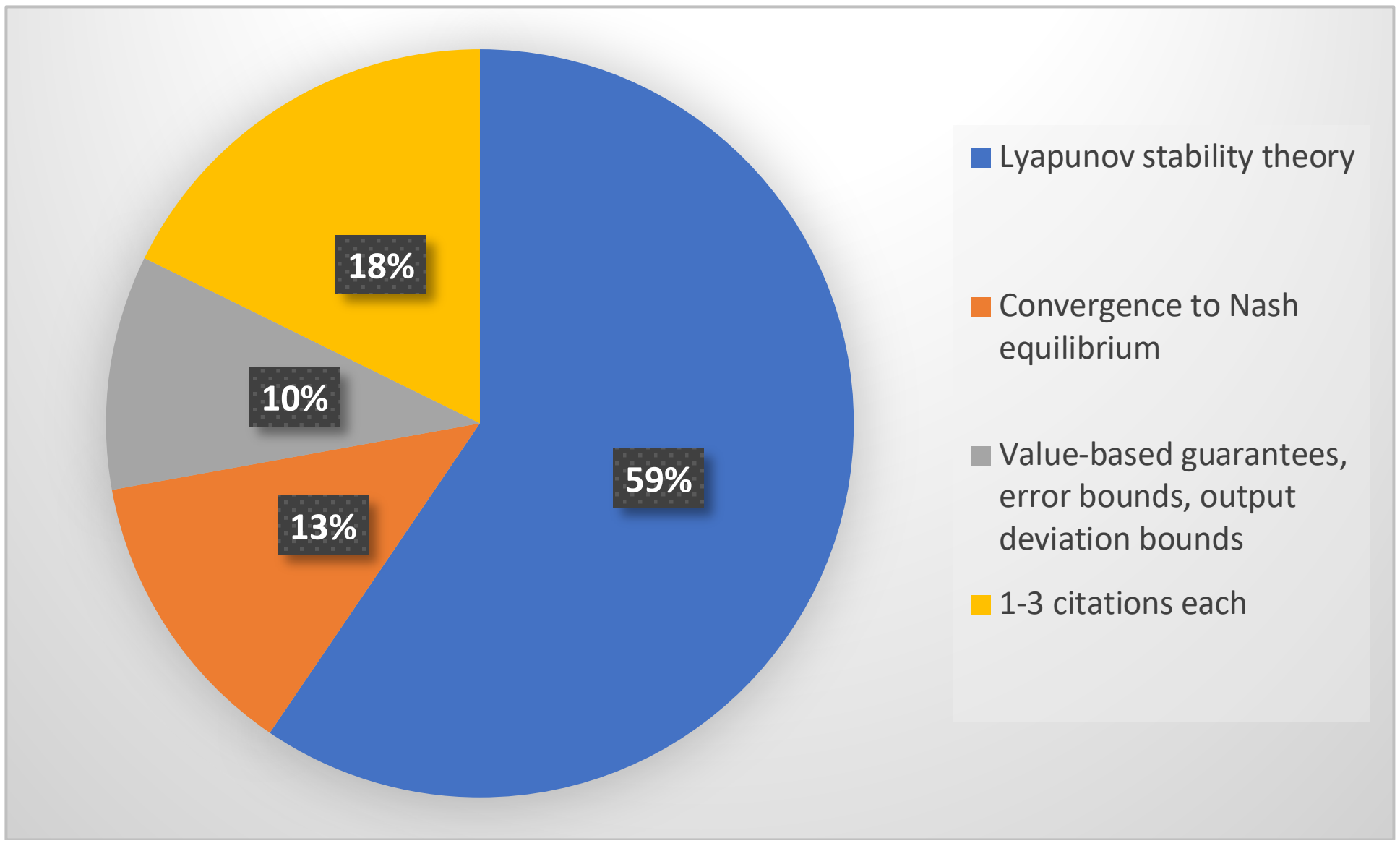

**Figure 7.** Theoretical Approaches

#### *3.3.6 Theoretical Approach Objectives*

We also reviewed the *<behavior> objective* for theoretical papers (Table 11). Most ($n$=42, 54.5%) papers on theoretical approaches did not state their objective. Of those few that did, changes or dynamics in the operational environment was the most frequent objective ($n$=10), followed by differences or uncertainties in model parameters ($n$=7), adversarial attack ($n$=6), error or failure ($n$=5), differences between training and test or operational environments ($n$=2), input variation ($n$=2), then 1 each for domain shifts, different function approximation architectures and differences in quantization levels.

## 4. Discussion

Our study was conducted to characterize published means of measuring or determining the stability, robustness or resilience of RL. Out of an initial collection of 16,015 items 248 papers met inclusion criteria and were systematically reviewed. Approaches to measuring or determining the behavior were classified as either quantitative or theoretical. Quantitative approaches were further classified as internal or external depending on whether they evaluated the training phase or the test or operational phases. For both categories of quantitative approaches, we categorized the metrics used, with internal approaches primarily using the reward or score (and statistics on same) and external approaches primarily using variations on performance-related metrics

**Table 11.** Theoretical <*behavior*> objectives

| <*behavior*> Objective | Behavior(s) | Citations | Total |
|---|---|---|---|
| Not specifically stated | Stability<br>Robustness | [4, 5, 10, 15, 16, 20, 22, 24, 25, 33, 37, 39, 41, 43, 44, 46, 47, 48, 49, 52, 54, 56, 57, 60, 61, 62, 63, 64, 65, 67, 74, 76, 78, 80]<br>[90, 105, 116, 121, 165, 184, 197, 200, 216, 227, 240] | 42 |
| Changes in the operational environment/dynamic environment, environment uncertainties, environment disturbances | Stability<br>Robustness | [35, 79]<br>[135, 185, 194, 195, 222, 225, 228, 248] | 10 |
| Differences/uncertainties in model parameters | Stability<br>Robustness | [36, 55, 68]<br>[91, 201, 205, 225] | 7 |
| Adversarial attack, corruption, perturbations | Robustness | [150, 206, 215, 223, 230, 239] | 6 |
| Error, failure | Stability<br>Robustness | [69, 71]<br>[135, 145, 202] | 5 |
| Differences between training and test or operational environments | Stability<br>Robustness | [40]<br>[215] | 2 |
| Input variation or perturbation | Robustness | [195, 226] | 2 |
| Domain shifts | Robustness | [229] | 1 |
| Function approximation architecture | Robustness | [233] | 1 |
| Quantization level | Robustness | [250] | 1 |
| | | **Total** | **77** |

If a paper used multiple theoretical approaches, it was placed in each of those categories.

(though not precision, accuracy or recall). Theoretical approaches were dominated by the use of Lyapunov stability theory. We further characterized the objectives of the stability, robustness and resilience behaviors. Quantitative approaches to measuring the behavior focused on the ability to handle differences in the operational environment, whereas the vast majority of theoretical approaches to determining the behavior did not specifically state an objective. However, the objective of the theoretical approaches can be implied by the use of Lyapunov stability theory, that is, to prove the stability of the system. Lyapunov was used regardless of whether the article was on stability or robustness.

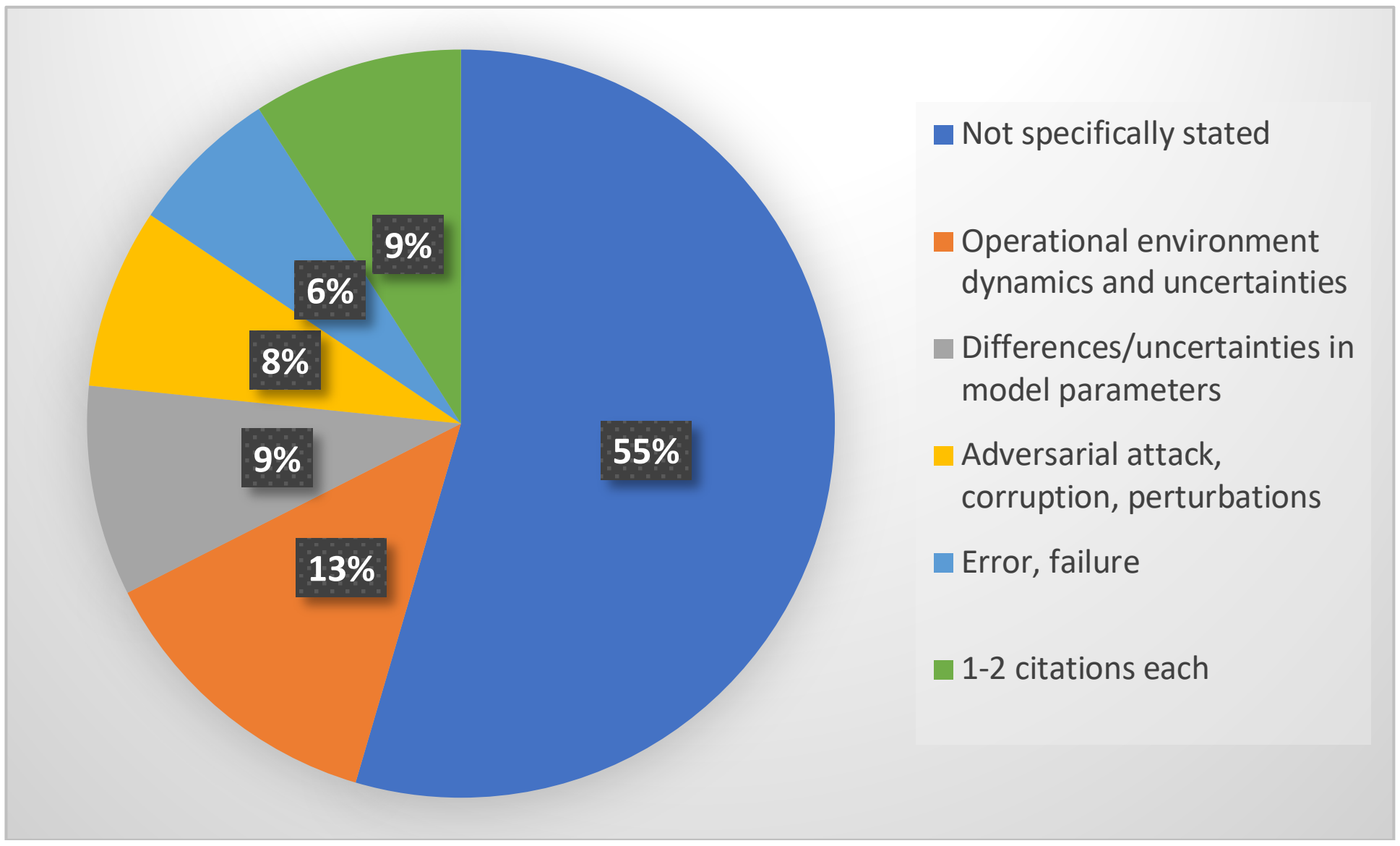

**Figure 8.** Theoretical <*behavior*> objectives

As an aid to deciding which metric to use, we developed a decision tree based on the information obtained in this literature review. It is a collapsible tree so that branches are not exposed unless selected and open branches can be closed or collapsed. There are several levels in the decision tree, starting with the i) behavior (stability, robustness or resilience); ii) the domain (see Table 3); iii) a list of quantitative and theoretical objectives (see Tables 9, 11); iv) the next level divides the metrics into external, internal and theoretical metrics (see Tables 5, 6); and v) the last level, i.e., the leaves, is the set of metrics for that branch of the decision tree (see Tables 7, 8). For example, suppose we want to find a suitable metric to measure robustness of a control system that is expected to face changes in the operational environment. From the metric decision tree shown in Figure 14, we see that the first selection is for a robustness metric. This selection displays the domains in which robustness metrics were described. Selecting the General Control domain reveals 5 quantitative objectives and 4 theoretical objectives, including the objective “Changes in the Operational Environment” in both the quantitative and theoretical objectives. An external metric found in the literature for this case is “blood glucose response” which is not applicable for this control system. The more appropriate metrics and approaches are Lyapunov stability theory and calculation, size of the stability region and value-based guarantees. One or all of these can be used to measure robustness of a general control system to changes in the operational environment.

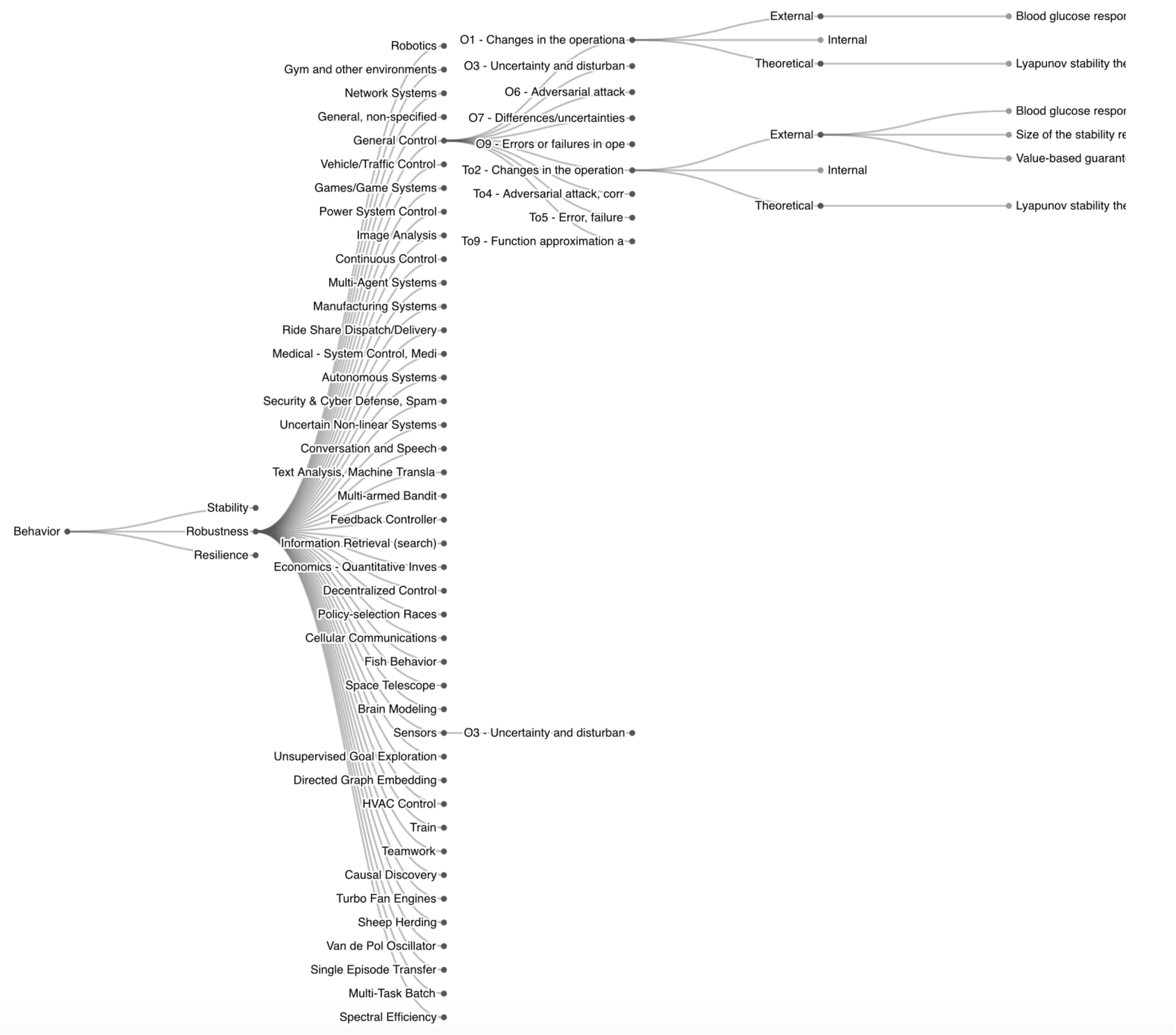

**Figure 9**. Metric Selection Decision Tree

## 5. Supporting Information

The databases in Table 12 were readily available to the author through Oak Ridge National Laboratory library subscriptions. Databases selected for this study are international in scope and ensure we investigate relevant studies that are global and cover a wide range of application domains, thereby eliminating the risk of bias from the author subject matter expertise and a Western perspective. However, we are bound by the content of these databases.

Many abstracting and indexing databases have very broad coverage, which results in individual databases duplicating content found in others. Duplicated bibliographic records were identified and removed. After duplicate entries were removed, the remaining citations were reviewed for completeness in terms of information needed to obtain the document from a library. If the citation did not include this information (e.g., author name(s), article title, and journal name), the missing information was obtained from the source database or other online sources and the citation was manually corrected.

**Table 12.** Information sources used in study

| **Database** | **Dates of Coverage** | **Type of Database** |
|---|---|---|
| arXiv | 1991-present | Publicly Available / Open Access |
| Scopus | 1823-present | Subscription |
| Web of Science | 1900-present | Subscription |

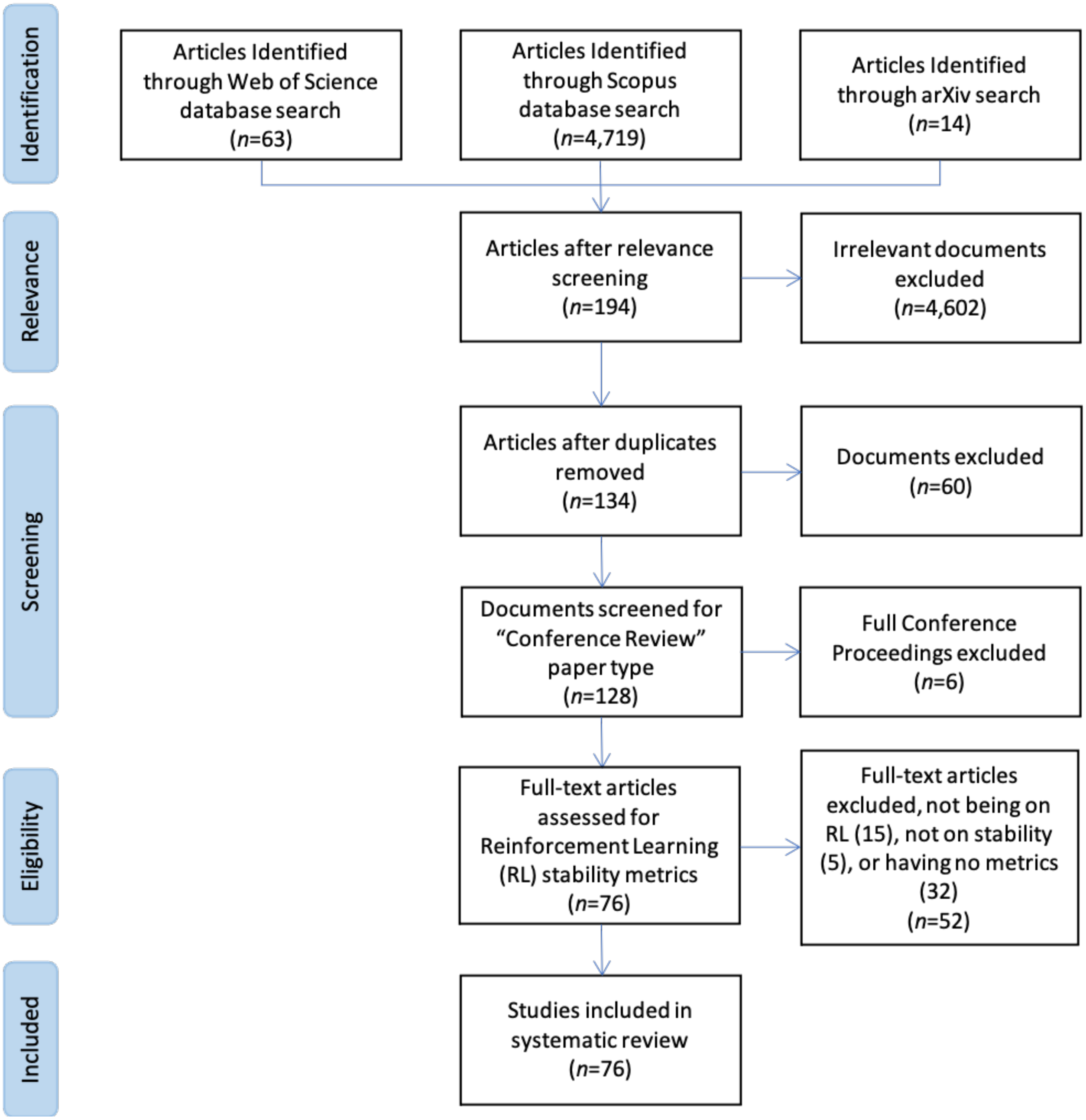

**Figure 10.** The PRISMA Flow Diagram for Stability

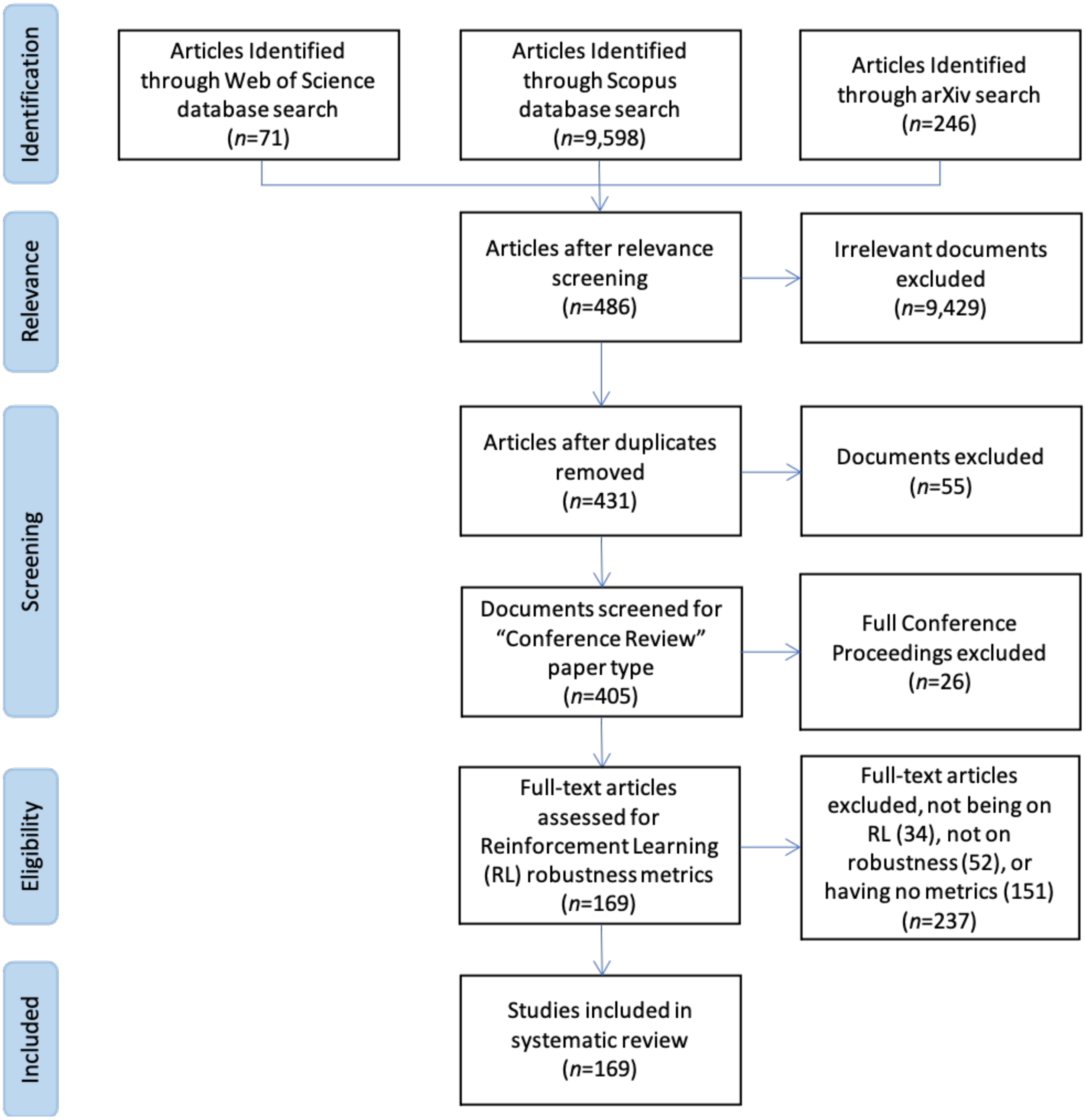

**Figure 11.** The PRISMA Flow Diagram for Robustness

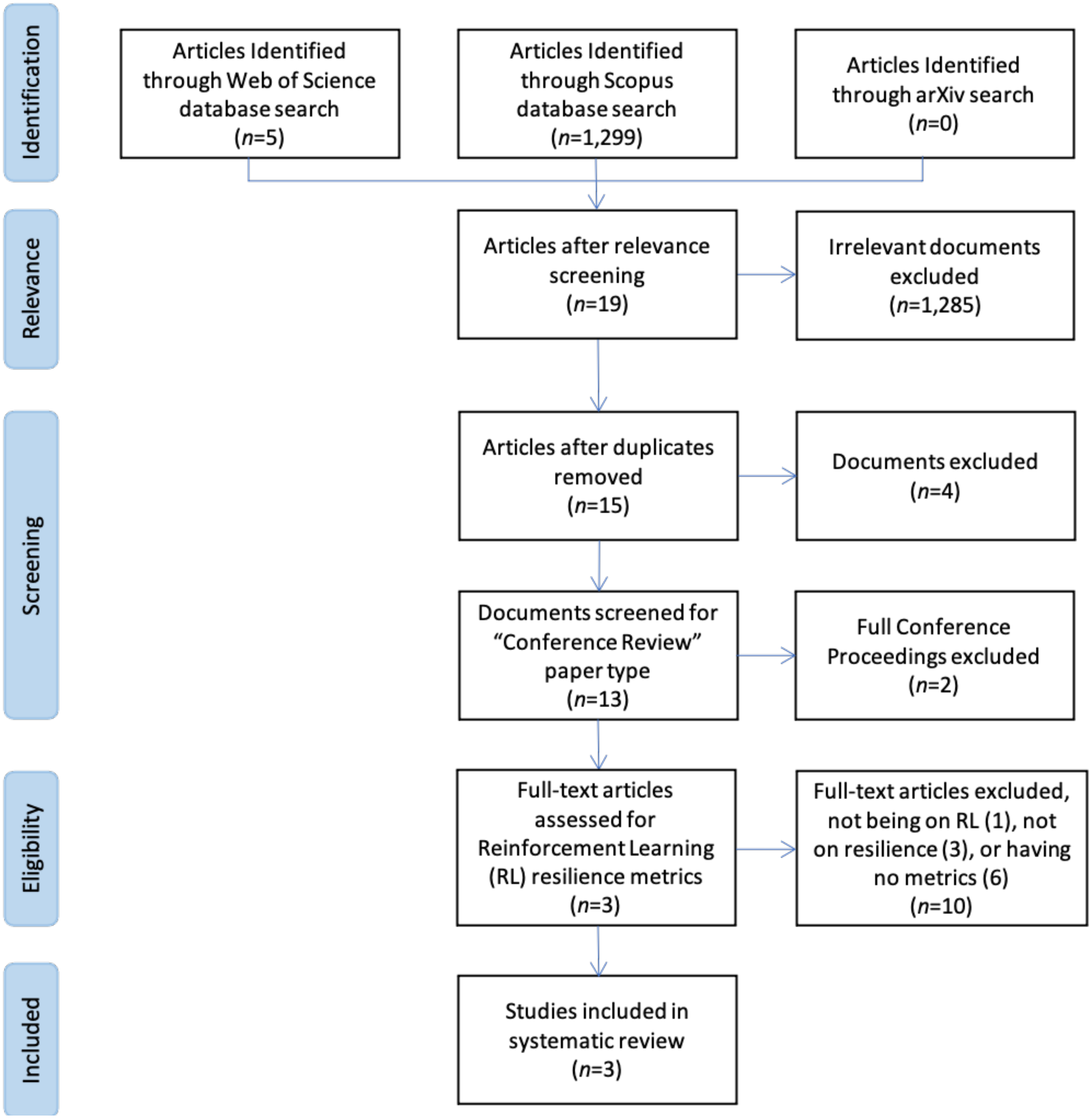

**Figure 12.** The PRISMA Flow Diagram for Resilience

**Table 13.** Data Reduction Summary

| Search phrases | "reinforcement learning" AND stability AND ("metric" OR "measure" OR "index" OR "score" OR "quantifier" OR "indicator") [in Title, Abstract, & Keywords] | | | Search phrases | "reinforcement learning" AND robust* AND ("metric" OR "measure" OR "index" OR "score" OR "quantifier" OR "indicator") [in Title, Abstract, & Keywords] | | | Search phrases | "reinforcement learning" AND resilien* AND ("metric" OR "measure" OR "index" OR "score" OR "quantifier" OR "indicator") [in Title, Abstract, & Keywords] | | |
|---|---|---|---|---|---|---|---|---|---|---|---|
| Topic | Stability | | | Topic | Robustness | | | Topic | Resilience | | |
| Search DB | WoS | Scopus | arXiv | Search DB | WoS | Scopus | arXiv | Search DB | WoS | Scopus | arXiv |
| Original Count | 63 | 4,719 | 14 | Original Count | 71 | 9,598 | 246 | Original Count | 5 | 1,299 | 0 |
| in TAK | 118 | | | in TAK | 171 | | | in TAK | 14 | | |
| Duplicate in WoS | -- | 6 | 0 | Duplicate in WoS | -- | 7 | 0 | Duplicate in WoS | -- | -- | -- |
| Duplicate in Scopus | 49 | 5 | 0 | Duplicate in Scopus | 46 | 1 | 1 | Duplicate in Scopus | 4 | -- | -- |
| Duplicate in arXiv | 0 | 0 | -- | Duplicate in arXiv | -- | -- | -- | Duplicate in arXiv | 0 | -- | -- |
| Books | 0 | 0 | 0 | Books | 0 | 0 | 0 | Books | 0 | 0 | -- |
| Editorials | 0 | 0 | 0 | Editorials | 0 | 0 | 0 | Editorials | 0 | 0 | -- |
| No paper, no abs | 0 | 1 | 0 | No paper, no abs | 0 | 1 | 1 | No paper, no abs | 0 | 0 | 0 |
| Conference Reviews | 0 | 6 | 0 | Conference Reviews | 0 | 26 | 0 | Conference Reviews | 0 | 2 | |
| Subtotal | 14 | 100 | 14 | Subtotal | 25 | 136 | 244 | Subtotal | 1 | 12 | 0 |
| Inverse RL Papers or Not RL | 0 | 14 | 1 | Inverse RL Papers or Not RL | 1 | 13 | 20 | Inverse RL Papers or Not RL | 1 | 0 | -- |
| Not Stability | 1 | 3 | 1 | Not Robustness | 3 | 41 | 7 | Not Resilience | -- | 3 | -- |
| Papers with no Metrics or Theory | 0 | 29 | 3 | Papers with no Metrics | 14 | 37 | 100 | Papers with no Metrics | -- | 6 | -- |
| Subtotal | 13 | 54 | 9 | Subtotal | 7 | 45 | 117 | Subtotal | 0 | 3 | 0 |
| Specific Metric | 2 | 3 | 1 | Specific Metric | 0 | 0 | 2 | Specific Metric | 0 | 0 | 0 |
| Performance Related Metric | 4 | 22 | 3 | Performance Related Metric | 5 | 37 | 88 | Performance Related Metric | 0 | 3 | 0 |
| Theoretical Approach | 5 | 26 | 3 | Theoretical Approach | 2 | 4 | 6 | Theoretical Approach | 0 | 0 | 0 |
| Both Perf and Theo | 2 | 4 | 1 | Both Perf and Theo | 0 | 2 | 17 | Both Perf and Theo | 0 | 0 | 0 |
| Both Stab & Theo | 0 | 0 | 0 | Both Rob & Theo | 0 | 1 | 3 | Both Res & Theo | 0 | 0 | 0 |
| Both Stab & Perf metric | 0 | 0 | 0 | Both Rob & Perf metric | 0 | 1 | 0 | Both Res & Perf metric | 0 | 0 | 0 |
| Stab & Perf Metric & Theeory | 0 | 0 | 0 | Rob & Perf Metric & Theeory | 0 | 0 | 1 | Res & Perf Metric & Theeory | 0 | 0 | 0 |

**Table 13.** The PRISMA Checklist [251]

| Section/topic | # | Checklist item | Reported in Section |
|---|---|---|---|
| **TITLE** | | | |
| Title | 1 | Identify the report as a systematic review, meta-analysis, or both. | Introduction |
| **ABSTRACT** | | | |
| Structured summary | 2 | Provide a structured summary including, as applicable: background; objectives; data sources; study eligibility criteria, participants, and interventions; study appraisal and synthesis methods; results; limitations; conclusions and implications of key findings; systematic review registration number. | Introduction |
| **INTRODUCTION** | | | |
| Rationale | 3 | Describe the rationale for the review in the context of what is already known. | Introduction |
| Objectives | 4 | Provide an explicit statement of questions being addressed with reference to participants, interventions, comparisons, outcomes, and study design (PICOS). | Introduction |
| **METHODS** | | | |
| Protocol and registration | 5 | Indicate if a review protocol exists, if and where it can be accessed (e.g., Web address), and, if available, provide registration information including registration number. | Introduction |
| Eligibility criteria | 6 | Specify study characteristics (e.g., PICOS, length of follow-up) and report characteristics (e.g., years considered, language, publication status) used as criteria for eligibility, giving rationale. | Methods and Supporting Information (SI) |
| Information sources | 7 | Describe all information sources (e.g., databases with dates of coverage, contact with study authors to identify additional studies) in the search and date last searched. | Methods and SI |
| Search | 8 | Present full electronic search strategy for at least one database, including any limits used, such that it could be repeated. | SI |
| Study selection | 9 | State the process for selecting studies (i.e., screening, eligibility, included in systematic review, and, if applicable, included in the meta-analysis). | Methods |

**Table 13.** The PRISMA Checklist [251]

| Section/topic | # | Checklist item | Reported in Section |
|---|---|---|---|
| **METHODS (Continued)** | | | |
| Data collection process | 10 | Describe method of data extraction from reports (e.g., piloted forms, independently, in duplicate) and any processes for obtaining and confirming data from investigators. | Methods |
| Risk of bias in individual studies | 12 | Describe methods used for assessing risk of bias of individual studies (including specification of whether this was done at the study or outcome level), and how this information is to be used in any data synthesis. | Methods and SI |
| Summary measures | 13 | State the principal summary measures (e.g., risk ratio, difference in means). | Methods |
| Synthesis of results | 14 | Describe the methods of handling data and combining results of studies, if done, including measures of consistency (e.g., $I^2$) for each meta-analysis. | Methods |
| Risk of bias across studies | 15 | Specify any assessment of risk of bias that may affect the cumulative evidence (e.g., publication bias, selective reporting within studies). | Methods |
| Additional analyses | 16 | Describe methods of additional analyses (e.g., sensitivity or subgroup analyses, meta-regression), if done, indicating which were pre-specified. | Methods |
| **RESULTS** | | | |
| Study selection | 17 | Give numbers of studies screened, assessed for eligibility, and included in the review, with reasons for exclusions at each stage, ideally with a flow diagram. | Results and SI |
| Study characteristics | 18 | For each study, present characteristics for which data were extracted (e.g., study size, PICOS, follow-up period) and provide the citations. | Results |
| Risk of bias within studies | 19 | Present data on risk of bias of each study and, if available, any outcome level assessment (see item 12). | Results |
| Results of individual studies | 20 | For all outcomes considered (benefits or harms), present, for each study: (a) simple summary data for each intervention group (b) effect estimates and confidence intervals, ideally with a forest plot. | Results |
| Synthesis of results | 21 | Present results of each meta-analysis done, including confidence intervals and measures of consistency. | Results |
| Risk of bias across studies | 22 | Present results of any assessment of risk of bias across studies (see Item 15). | Results |

**Table 13.** The PRISMA Checklist [251]

| Section/topic | # | Checklist item | Reported in Section |
|---|---|---|---|
| **RESULTS (Continued)** | | | |
| Additional analysis | 23 | Give results of additional analyses, if done (e.g., sensitivity or subgroup analyses, meta-regression [see Item 16]). | Results |
| Data items | 11 | List and define all variables for which data were sought (e.g., PICOS, funding sources) and any assumptions and simplifications made. | Methods |
| **Section/topic** | **#** | **Checklist item** | **Reported in Section** |
| **DISCUSSION** | | | |
| Summary of evidence | 24 | Summarize the main findings including the strength of evidence for each main outcome; consider their relevance to key groups (e.g., healthcare providers, users, and policy makers). | Discussion |
| Limitations | 25 | Discuss limitations at study and outcome level (e.g., risk of bias), and at review-level (e.g., incomplete retrieval of identified research, reporting bias). | Discussion |
| Conclusions | 26 | Provide a general interpretation of the results in the context of other evidence, and implications for future research. | Discussion |
| **FUNDING** | | | |
| Funding | 27 | Describe sources of funding for the systematic review and other support (e.g., supply of data); role of funders for the systematic review. | Acknowledgements |

## 6. Acknowledgements

The author would like to acknowledge Rama Vasudevan, PhD of Oak Ridge National Laboratory for intellectual discussions and collaborative research on reinforcement learning. The author would also like to thank Nathan Martindale for assistance making the decision tree more readable and thus usable.

This work was funded initially by the AI Initiative at the Oak Ridge National Laboratory and subsequently funded by the US Department of Energy, National Nuclear Security Administration's Office of Defense Nuclear Nonproliferation Research and Development (NA-22).

**Acronyms**

| | |
|---|---|
| A2C | Advantage Actor Critic |
| AODV | Ad-hoc On-demand Distance Vector |
| ARVI | Anytime Reduced Value Iteration |
| ATTN | Active Tracking Target Network |
| AUC | Area Under the receiver operating characteristic (ROC) Curve (AUC) |
| BMI | Brain Machine Interface |
| CLAC | Lyapunov-based Actor Critic |
| CPPO | Constraint-controlled Proximal Policy Optimization |
| DB | Database |
| DDPG | Deep Deterministic Policy Gradient |
| DOE | Department of Energy |
| DrAC | Data-regulated Actor-Critic |
| HVAC | Heating, Ventilation, and Air Conditioning |
| JSD | Jensen-Shannon Divergence |
| Max | Maximum |
| Min | Minimum |
| MOOSE | Model-based Offline policy Search with Ensembles |
| NESM | Normalized Energy Stability Margin |
| Perf | Performance |
| PPO | Proximal Policy Optimization |
| PRISMA | Preferred Reporting Items for Systematic Reviews and Meta-Analyses |
| RL | Reinforcement Learning |
| Res | Research |
| Rob | Robustness |
| ROC | Receiver Operating Characteristic |
| sec | seconds |
| Stab | Stability |
| SVD | Singular Value Decomposition |
| TAK | Title, Abstract and Keywords |
| Theo | Theoretical |
| US | United States |
| WoS | Web of Science |

**References**

*Resilience*

*Stability*

*Robustness*

*Other*